%% file: main.tex
\documentclass[letterpaper]{article} % DO NOT CHANGE THIS
\usepackage{aaai2026}  % DO NOT CHANGE THIS
\usepackage{times}  % DO NOT CHANGE THIS
\usepackage{helvet}  % DO NOT CHANGE THIS
\usepackage{courier}  % DO NOT CHANGE THIS
\usepackage[hyphens]{url}  % DO NOT CHANGE THIS
\usepackage{graphicx} % DO NOT CHANGE THIS
\usepackage{natbib}  % DO NOT CHANGE THIS AND DO NOT ADD ANY OPTIONS TO IT
\usepackage{caption} % DO NOT CHANGE THIS AND DO NOT ADD ANY OPTIONS TO IT
\usepackage{algorithm}
\usepackage{algorithmic}
\usepackage{amsmath} 
\usepackage{amsfonts}
\usepackage{subcaption}
\usepackage{multirow, adjustbox}
\usepackage{listings}
\usepackage{booktabs}
\usepackage{xr}
\usepackage{cuted}
\usepackage{placeins}

\usepackage{newfloat}
\usepackage{listings}
\DeclareCaptionStyle{ruled}{labelfont=normalfont,labelsep=colon,strut=off} % DO NOT CHANGE THIS
\floatstyle{ruled}
\newfloat{listing}{tb}{lst}{}
\floatname{listing}{Listing}
\usepackage{float}
\nocopyright
\title{T-LoRA: Single Image Diffusion Model Customization Without Overfitting}
\author{
    Vera Soboleva\textsuperscript{\rm 1, \rm 2}, \ Aibek Alanov\textsuperscript{\rm 1, \rm 2}, Andrey Kuznetsov\textsuperscript{\rm 1, \rm 3}, \ Konstantin Sobolev\textsuperscript{\rm 1, \rm 4}\thanks{Corresponding author: sobolevkv@my.msu.ru} \\
}
\affiliations{
    \textsuperscript{\rm 1}FusionBrain Lab, \
    \textsuperscript{\rm 2}HSE University, \
    \textsuperscript{\rm 3}Innopolis University, \
    \textsuperscript{\rm 4}Lomonosov Moscow State University
}

\usepackage{bibentry}
\usepackage{cuted}
\begin{document}

\maketitle
\begin{strip}
\vspace{-7em}
  \centering
  \includegraphics[width=\textwidth]{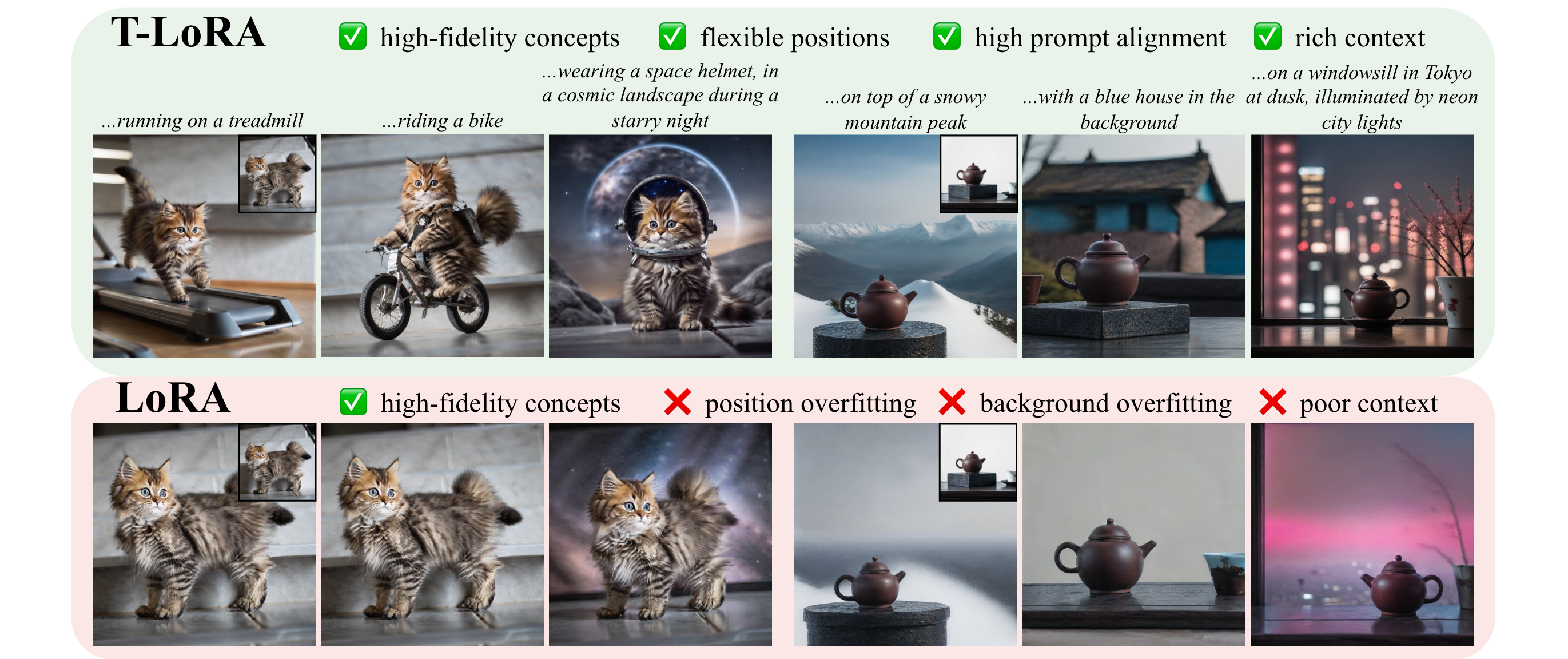}
  \captionof{figure}{\textit{T-LoRA} reduces overfitting related to position and background, enabling versatile and enriched generation.}
  \label{fig:visual_abs}
\end{strip}

% \begin{FullWidth}
% \twocolumn[{
% \begin{figure}[h]
% \includegraphics[width=\textwidth]{images/visual_abs.png}
% \caption[width=\textwidth]{Compared to standard LoRA single view fine-tuning, \textit{T-LoRA} reduces overfitting related to position and background, enabling more versatile and enriched generation.}
% \label{fig:visual_abs}
% \end{figure}
% \bigskip
% }]
% \end{FullWidth}

\input{sections/00_abstract}

\input{sections/01_intro}

\input{sections/02_background}
\input{sections/04_method_arxiv}
\input{sections/05_experiments_arxiv}
\input{sections/06_conclusion}

% \newpage
% \pagebreak

% \bibliographystyle{plain}
% \bibliographystyle{plainnat}
% \bibliographystyle{neurips_2025}
\bibliography{aaai2026}

\appendix
\input{sections/appendix_extended}
\newpage

% \input{sections/appendix}

% \bibliography{aaai2026}


\end{document}

%% file: sections/00_abstract.tex
\begin{abstract}

While diffusion model fine-tuning offers a powerful approach for customizing pre-trained models to generate specific objects, it frequently suffers from overfitting when training samples are limited, compromising both generalization capability and output diversity. This paper tackles the challenging yet most impactful task of adapting a diffusion model using just a single concept image, as single-image customization holds the greatest practical potential. We introduce \textit{T-LoRA}, a \textbf{T}imestep-Dependent \textbf{Lo}w-\textbf{R}ank \textbf{A}daptation framework specifically designed for diffusion model personalization. We show that higher diffusion timesteps are more prone to overfitting than lower ones, necessitating a timestep-sensitive fine-tuning strategy. \textit{T-LoRA} incorporates two key innovations: (1) a dynamic fine-tuning strategy that adjusts rank-constrained updates based on diffusion timesteps, and (2) a weight parametrization technique that ensures independence between adapter components through orthogonal initialization. Extensive experiments on SD-XL and FLUX-1.dev show that \textit{T-LoRA} and its individual components outperform standard LoRA and other diffusion model personalization techniques, achieving a superior balance between concept fidelity and text alignment.
\end{abstract}

\begin{links}
    \link{Project Page}{https://controlgenai.github.io/T-LoRA/}
\end{links}

%% file: sections/01_intro.tex
\section{Introduction} ~\label{sec:intro}
Recent advances in personalizing large-scale text-to-image diffusion models~\cite{DB, TI, CD} have revolutionized content creation, enabling pre-trained models with the ability to generate highly specific and customized outputs, such as particular objects, styles, or domains. The primary objectives of such customization are twofold: (1) ensuring high-quality preservation of the target concept, and (2) achieving precise alignment between the generated image and the input text.

Fine-tuning based customization approaches are effective at producing high fidelity concept samples~\cite{DB, CD}. However, they encounter limitations due to the restricted dataset size, which hinder generalization and cause artifacts like background elements and pose information to "leak" into outputs, reducing the model's flexibility and creativity. Compared to standard fine-tuning, lightweight Low-Rank Adaptater (LoRA)~\cite{lora} offers significant advantages~\cite{biderman2024lora,Ryu2024githublora}: it significantly reduces the number of trainable parameters, making it suitable for resource-constrained settings, and it is less prone to overfitting, thereby better preserving the original model's generative capabilities.

Customizing diffusion models with a single concept image is particularly challenging due to the high risk of overfitting, even for lightweight methods~\cite{ortogonal, lora, svdiff} (Figure~\ref{fig:visual_abs}). However, this task holds significant practical value, as users often lack multiple images of their concepts with varied backgrounds, making single-image customization a key focus of our research.

We hypothesize that the root cause of overfitting lies in the fine-tuning process applied during the noisiest steps of the diffusion process. At these steps, the model is trained to recover the training images from heavily corrupted inputs, which inadvertently restricts its capacity to generate diverse and flexible scene structures. Simultaneously, these noisy steps are crucial for preserving the structural coherence and fine-grained details of the target concept. Our analysis reveals that omitting these noisy steps during fine-tuning results in a substantial loss of fidelity, underscoring the critical trade-off between maintaining concept precision and enabling generative diversity (Figure~\ref{fig:tlora_motivation}). 

Based on our analysis, we introduce \textit{T-LoRA}, a Timestep-Dependent Low-Rank Adaptation framework for diffusion model customization. \textit{T-LoRA} prioritizes training capacity for less noisy timesteps while reducing signals for noisier ones using a time-dependent masking strategy, \textit{Vanilla T-LoRA}, which restricts higher-rank LoRA components during noisier timesteps. Our further analysis reveals that standard LoRA adapters often exhibit an effective rank that is significantly smaller than the original rank hyperparameter. This property could limit the effectiveness of \textit{Vanilla T-LoRA}, as masked and unmasked adapters may possess similar expressive power. To address this, we introduce \textit{Ortho-LoRA}, a novel LoRA initialization technique ensuring orthogonality between adapter components, thereby explicitly separating information flows across different timesteps. Extensive experiments demonstrate the effectiveness of \textit{T-LoRA} and its components (\textit{Vanilla T-LoRA} and \textit{Ortho-LoRA}). Our proposed framework significantly outperforms existing lightweight fine-tuning approaches in single-image diffusion model personalization tasks in both metrics and user study. These results highlight the potential of integrating time-dependent regularization and orthogonality into future diffusion model adaptation frameworks.

To summarize, our key contributions are as follows:

\begin{itemize} 

\item We perform a detailed analysis of overfitting in diffusion model adaptation and reveal that it primarily occurs at higher (noisier) timesteps of the diffusion process.

\item We propose \textit{T-LoRA}, a Low-Rank Adaptation framework that mitigates overfitting in diffusion model personalization through a \textit{rank masking strategy}, balancing training capacity and applying stronger regularization at higher timesteps.

\item  We explore the concept of effective rank in LoRA matrices and propose \textit{Ortho-LoRA}, a novel \textit{orthogonal weight initialization method}, that enhances effective rank utilization and improves information flow separation across timesteps, boosting \textit{T-LoRA} performance.

\end{itemize}

%% file: sections/02_background.tex
\section{Background}

\textbf{Diffusion models}~\cite{song2020denoising, ho2020denoising} are probabilistic generative models using neural networks to approximate data distributions by iteratively denoising Gaussian-sampled variables. We focus on text-to-image diffusion models $\varepsilon_{\theta}$, which generate images $x$ from text prompt $P$. These models use a text encoder $E_T$ to extract text embeddings $p = E_T(P)$. Latent diffusion models~\cite{Rombach_2022_CVPR} encode images into latent representations $z = E(x)$ via an encoder $E$ and decode them with a decoder $D$, ensuring $x \approx D(z)$. The diffusion model $\varepsilon_{\theta}$ is trained to predict noise:

\begin{equation}\label{eq:training}
\min_\theta \mathbb{E}_{p, t, z, \varepsilon}\left[\left\|\varepsilon - \varepsilon_\theta(t, z_t, p)\right\|_2^2\right],
\end{equation}
\noindent
where $\varepsilon \sim N(0,I)$, $t$ is the diffusion timestep, and $z_t$ is a noisy latent code obtained via the forward process $z_t = \text{ForwardDiffusion}(t, z_0, \varepsilon)$. During inference, random noise  $z_T \sim N(0,I)$ is iteratively denoised to recover $z_0$.

\textbf{Low-Rank Adaptation (LoRA)}~\cite{lora} efficiently fine-tunes large pre-trained models by updating the weight matrix $W$ as $\tilde{W} = W + BA$, where $A \in \mathbb{R}^{r \times m}$, $B \in \mathbb{R}^{n \times r}$, $r$ is the rank, and $(n, m)$ is the dimensionality of $W$. To preserve the pre-trained model's behavior initially, $A$ is normally initialized, and $B$ is set to zero.

\textbf{Diffusion Model Customization} often involves fine-tuning, where model weights are adjusted to generate specific user-defined concepts from a limited set of concept images. To associate a new concept with a special text token $V^*$, $\varepsilon_{\theta}$ is fine-tuned on a small set of concept images $\mathbb{C} = \{x\}_{i=1}^N$ using the following objective:
\begin{equation}\label{eq:finetuning} \min_\theta \mathbb{E}_{p, t, z=\mathcal{E}(x), x \in \mathbb{C}, \varepsilon}\left[\left\|\varepsilon - \varepsilon_\theta(t, z_t, p)\right\|_2^2\right], \end{equation} 
\noindent

where $p = E_T(P)$ represents the text embeddings for the prompt $P =$ \textit{"a photo of a V*"}. Fine-tuning enhancements include pseudo-token optimization~\cite{TI}, diffusion model fine-tuning~\cite{DB, CD}, and lightweight parameterization~\cite{svdiff, ortogonal} to reduce costs and mitigate overfitting. LoRA~\cite{lora}, with its lightweight design, high concept fidelity, and strong prompt alignment, serves as a strong baseline for subject-driven generation and a key component in both fine-tuning-based~\cite{palp} and encoder-based~\cite{photomaker} personalization techniques.

\textbf{Overfitting Problem}  
Existing methods often overfit to position and background, especially with limited concept images. To address this, various techniques like image masking~\cite{elite,realcustom}, prompt augmentation~\cite{styledrop, tokenverse}, regularization~\cite{CD, core, attndreambooth}, advanced attention~\cite{realcustom,dreammatcher}, and sampling~\cite{profusion,beyondft} have been explored. In contrast, we identify the root cause as fine-tuning during noisiest timesteps, which reinforces background elements and limits flexibility. To address this, we propose reducing concept signals during noisy timesteps and adapting LoRA~\cite{lora} to control concept injection across timesteps, enhancing generalization and diversity while mitigating overfitting.

%% file: sections/04_method_arxiv.tex
\begin{figure*}[ht!]
  \centering
\includegraphics[width=\linewidth]{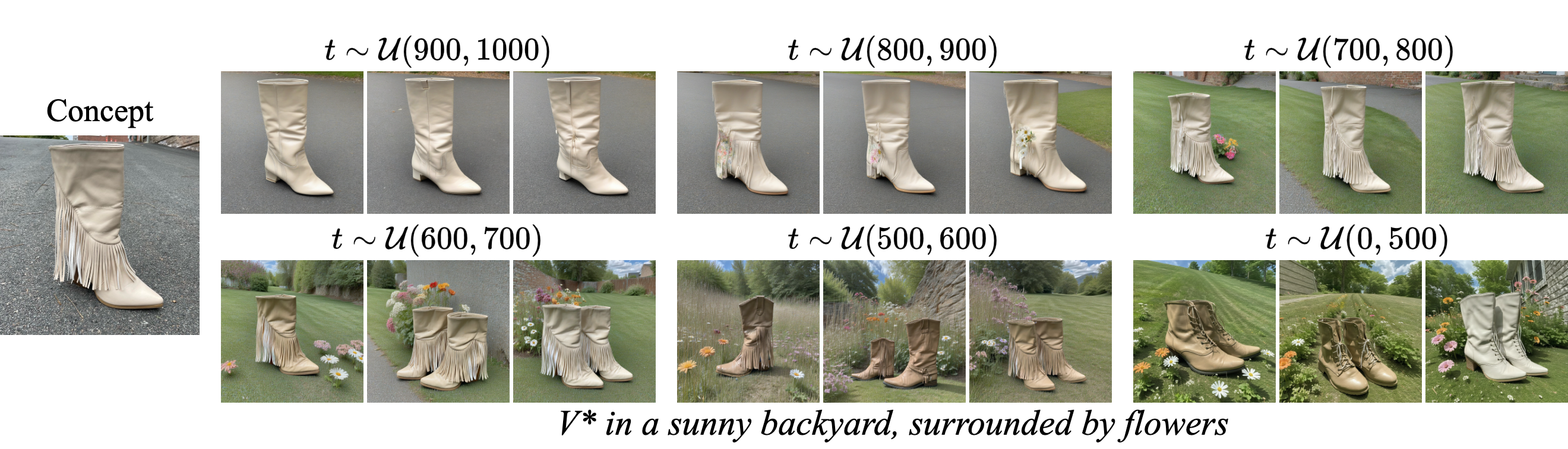}
\caption{\textbf{Motivational Experiment.} The figure shows fine-tuning the SD-XL model with LoRA over fixed timestep intervals. Focusing on the noisiest timesteps causes rapid overfitting, affecting positioning and backgrounds. Shifting to earlier timesteps improves text alignment and generation flexibility. However, completely excluding the noisiest timesteps ($t \in [800; 1000]$) is infeasible, as they are essential for maintaining concept fidelity.}

\label{fig:tlora_motivation}
\end{figure*}

\section{Method} \label{sec:method}

\textbf{Motivation}
Previous studies have shown that different timesteps in diffusion models play distinct roles throughout the generation process \cite{Choi2022CVPR, Koyejo2022NIPS, li2023autodiffusion}. For example, the authors of \cite{Choi2022CVPR} categorized the behavior of diffusion timesteps into three main stages: high timesteps  ($t \in [800; 1000]$) concentrate on forming coarse features, middle timesteps ($t \in [500; 800]$) produce perceptually rich content, and the lower timesteps ($t \in [0; 500]$) focus on removing residual noise. Additionally, works \cite{chang2023muse, gao2023mdtv2} show that high timesteps contribute to image diversity, as inadequate representation of the prompt's context during this stage makes it less likely to be restored in subsequent timesteps. Leveraging these insights, a growing body of work has introduced techniques to optimize diffusion model training. For instance, \cite{wang2025closer} proposed a time-dependent loss weighting strategy, while \cite{kim2025adaptive} and \cite{zheng2024beta} designed adaptive timestep sampling methods to improve the overall generation quality and diversity.

To investigate the role of different timesteps we fine-tuned the SD-XL model with LoRA over fixed timestep intervals (see Figure~\ref{fig:tlora_motivation}). The results show that fine-tuning at \textbf{higher timesteps} $t \in [800; 1000]$ leads to rapid overfitting, causing memorization of poses and backgrounds, which limits image diversity and prompt alignment, though these timesteps are crucial for defining shape and proportions.
In the \textbf{middle timesteps} $t \in [500; 800]$, the generated context became richer, and the model better reproduce fine concept details. However, we lose information related to the overall shape. For example, the boot in Figure~\ref{fig:tlora_motivation} retains fittings that correspond to the original concept, but the shoe is shorter. Finally, fine-tuning with \textbf{lower timesteps} $t \in [0; 500]$ demonstrated the best alignment with text prompts and yielded the richest generation. However, this approach struggled to accurately reproduce the intended concept, losing both shape and fine details of the object.

These findings highlight the necessity of managing the concept signal across timesteps. \textit{Concept information injection during noisier timesteps should be limited to encourage diversity. Middle timesteps should receive more concept information to produce correct fine details. The information at the lowest timesteps does not need to be restricted, as the risk of overfitting is minimal at this stage.}

\begin{figure*}[t]
\begin{center}

\begin{subfigure}{.32\linewidth}

\includegraphics[height=0.8\columnwidth,trim={1.5em 0em 1.5em 0em},clip]{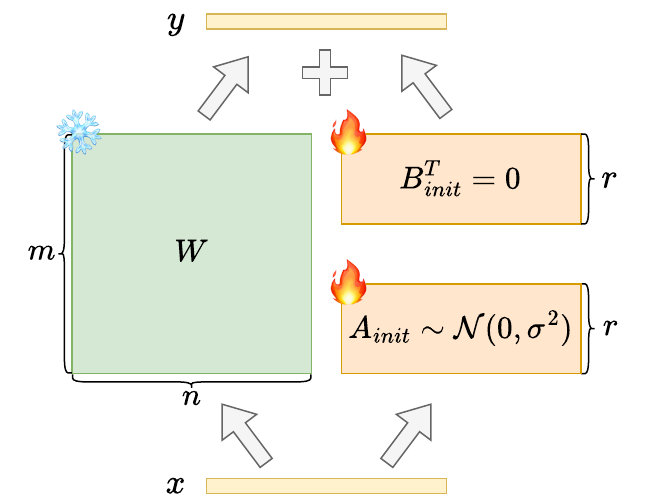}
\caption{LoRA}
\label{subfig:lora_scheme}
\end{subfigure}
\hfill
\begin{subfigure}{.32\linewidth}

\includegraphics[height=0.8\columnwidth,trim={1.5em 0em 1.5em 0em},clip]{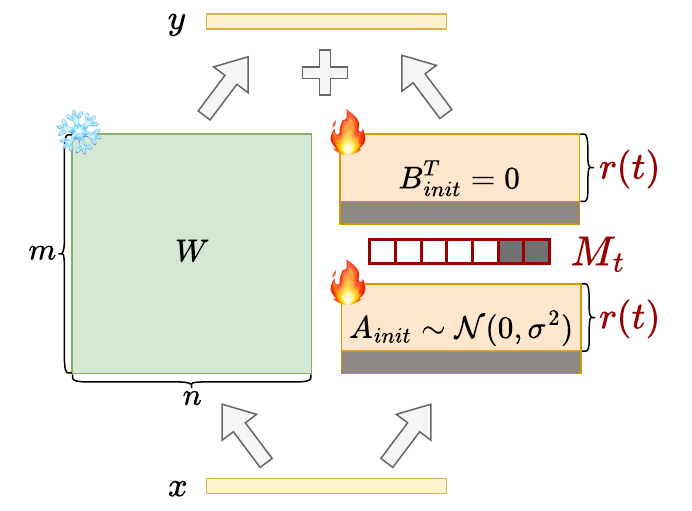}
\caption{Vanilla T-LoRA}
\label{subfig:vanilla_tlora_scheme}
\end{subfigure}
\hfill
\begin{subfigure}{.32\linewidth}

\includegraphics[height=0.8\columnwidth,trim={1.5em 0em 1.5em 0em},clip]{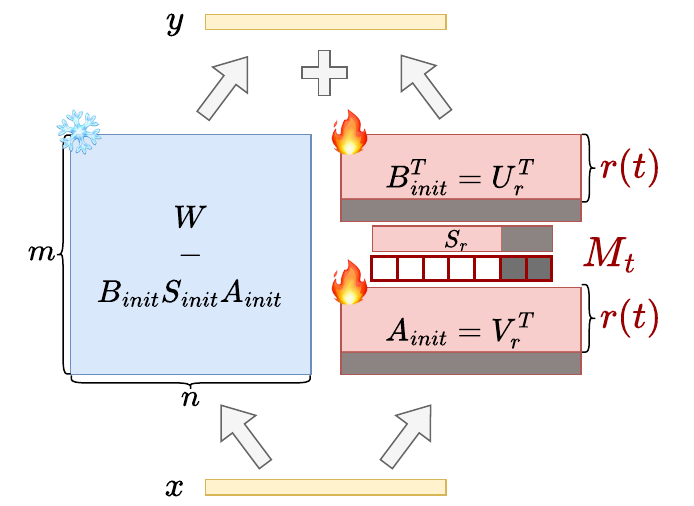}
\caption{T-LoRA}
\label{subfig:tlora_scheme}
\end{subfigure}

\caption{Comparison of training methods: LoRA, the proposed Vanilla T-LoRA, and T-LoRA schemes.}
\label{fig:method_scheme}

\end{center}
\end{figure*}

\subsection{Vanilla T-LoRA} \label{sec:tlora}

To tackle the aforementioned challenge, we propose a timestep-dependent fine-tuning strategy for diffusion models. Our method dynamically adjusts the ranks of LoRA adapters based on the diffusion timestep, allocating fewer parameters at higher timesteps and more at lower ones. This approach, called \textit{Vanilla T-LoRA}, incorporates a masking mechanism (see Figure~\ref{subfig:vanilla_tlora_scheme}):

\begin{equation} \label{eq:vanilla_tlora}
\centering
\begin{split}
    &\tilde{W}_{t} = W + B_{t} A_{t} = W + B M_{t} A, \ A \in \mathbb{R}^{r \times m}, B \in \mathbb{R}^{n \times r}
    \\
    &M_t = M_{r(t)} = \mathrm{diag}(\underbrace{1, 1, \dots, 1}_{r(t)}, \underbrace{0, 0, \dots, 0}_{r - r(t)}) \in \mathbb{R}^{r \times r}
\end{split}
\end{equation}

We define \( r(t) \) as a linear function inversely proportional to timesteps: \( r(t) = \lfloor (r - r_{\text{min}}) \cdot (T-t)/T \rfloor + r_{\text{min}} \), where \( r_{\text{min}} \) is a hyperparameter.  This rank-masking strategy dynamically controls information injection across timesteps during training and inference. Higher timesteps use smaller ranks to preserve generative capabilities by focusing on coarse features and context, while lower timesteps receive more information to capture fine concept details.

\subsection{On the LoRA Orthogonality} \label{sec:orthogonality}

\begin{figure}[t]
\begin{center}

\begin{subfigure}{\linewidth}
\centering
\includegraphics[width=0.49\linewidth]{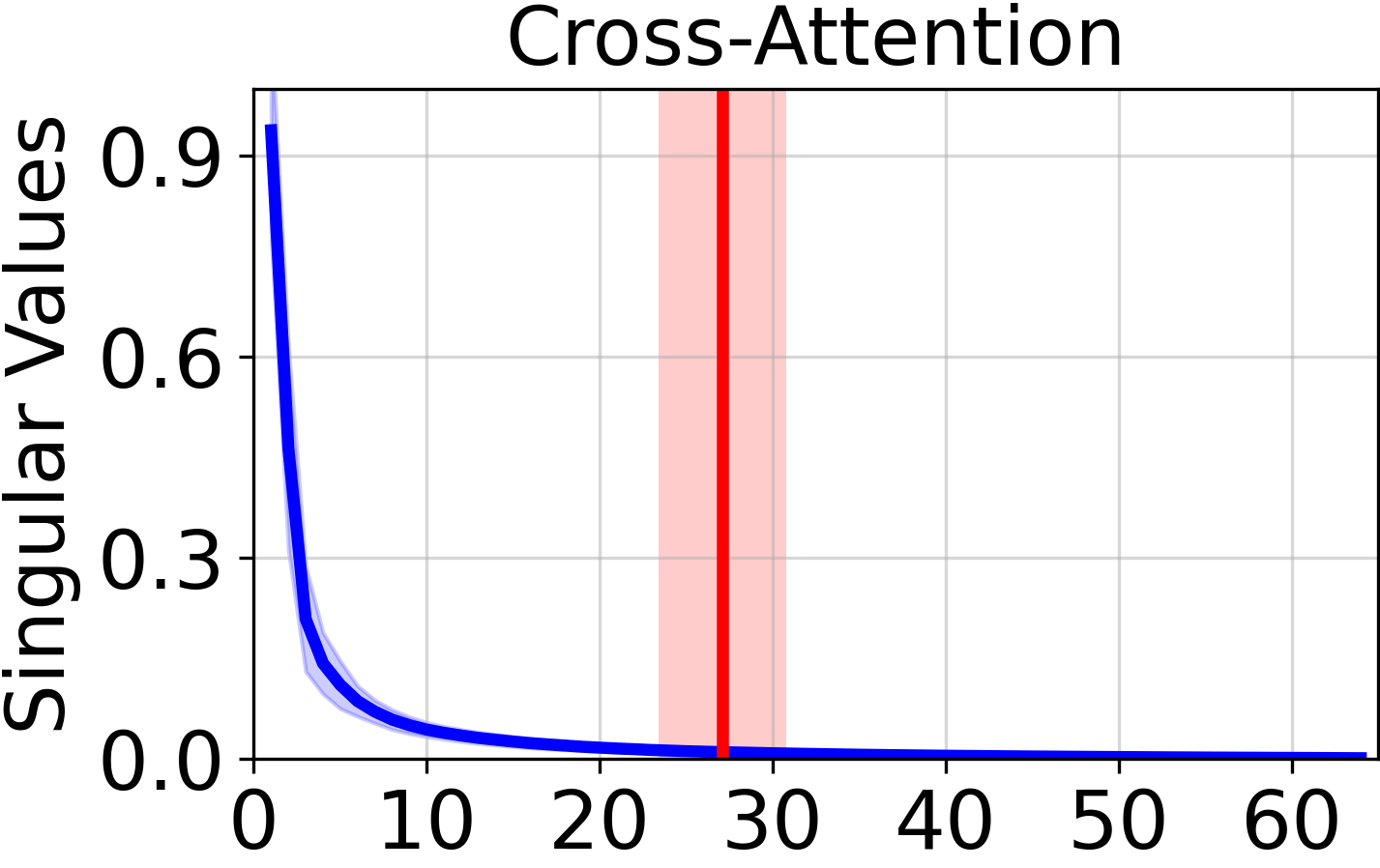}
\includegraphics[width=0.46\linewidth]{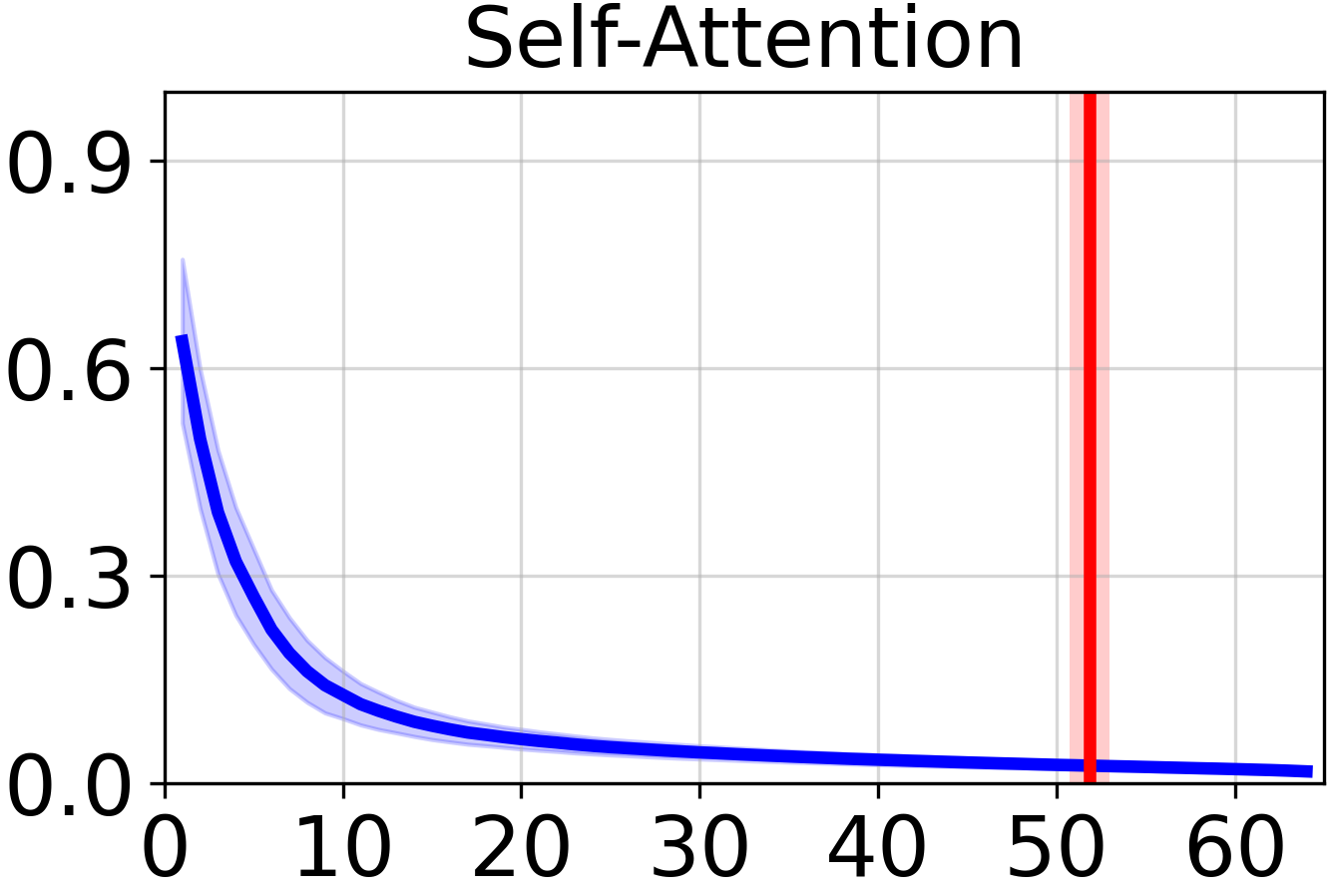}
\caption{LoRA}
\label{subfig:lora_sv}
\end{subfigure}
\vfill
\begin{subfigure}{\linewidth}
\centering
\includegraphics[width=0.49\linewidth]{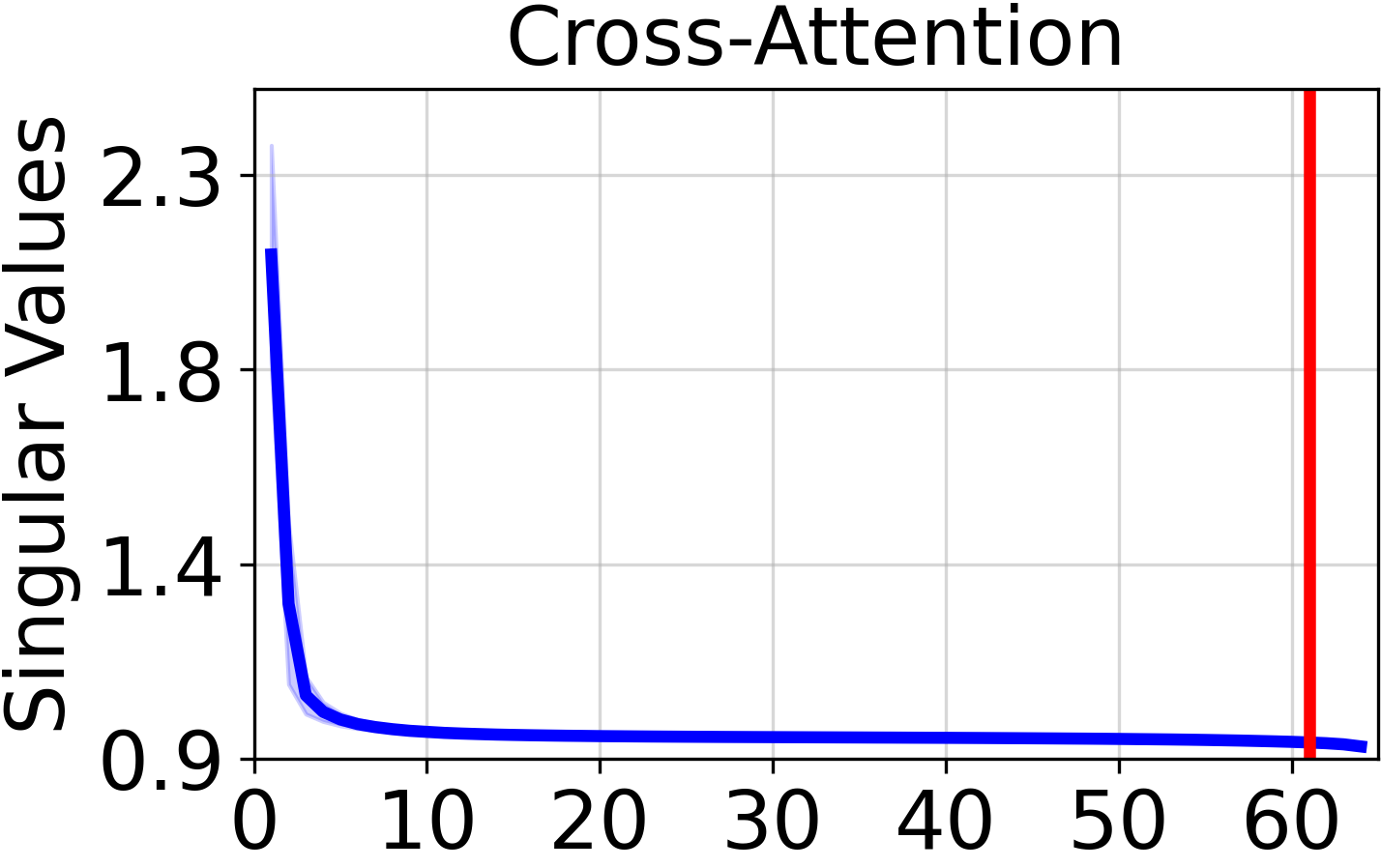}
\includegraphics[width=0.46\linewidth]{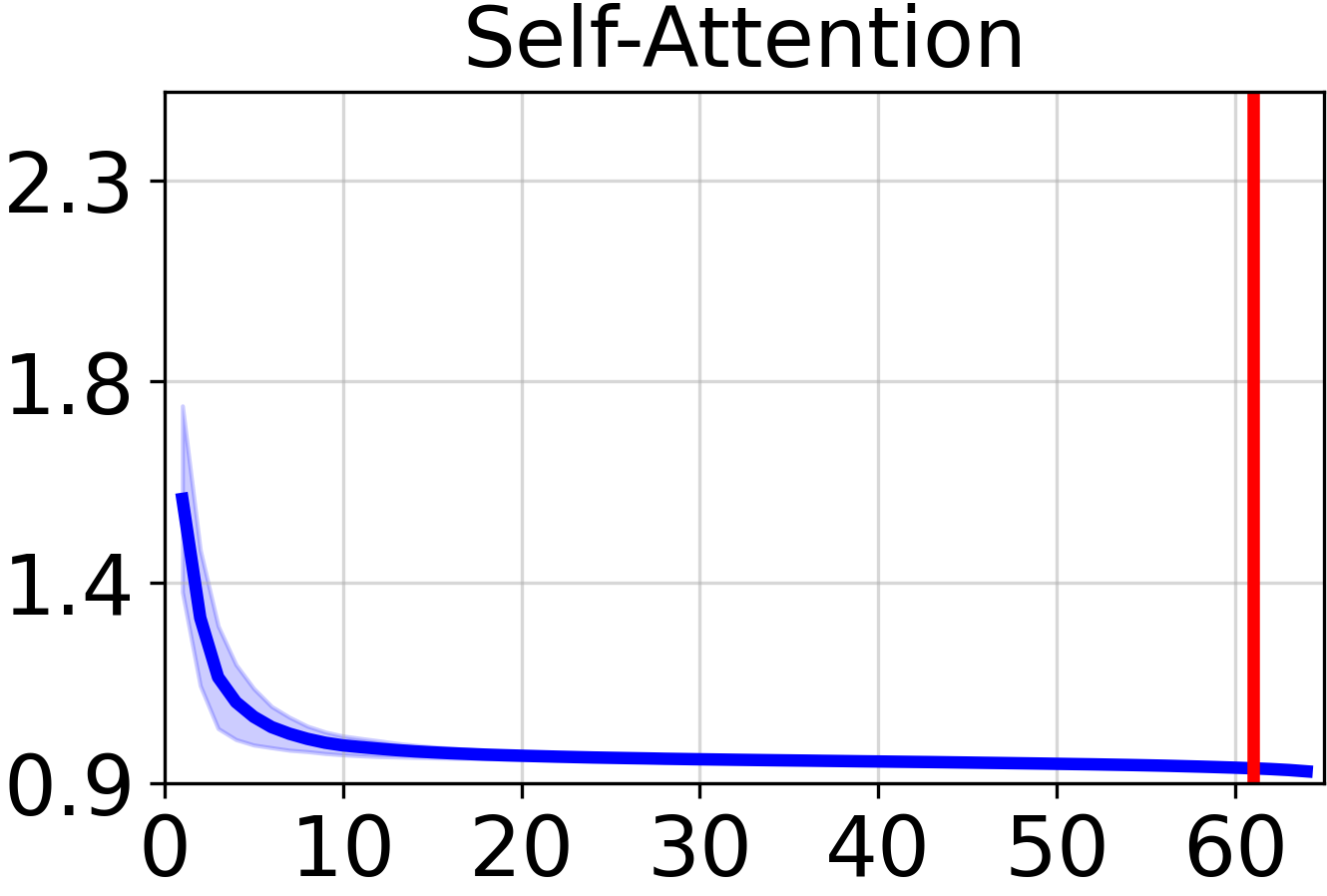}
\caption{Ortho-LoRA}
\label{subfig:ortholora_sv}
\end{subfigure}

\caption{Singular values of B matrices for LoRA and Ortho-LoRA with $r=64$ after 800 training steps. The red line marks the rank capturing \textbf{95\% of the total singular value sum}. LoRA matrices are effectively low-rank -- especially in cross-attention -- while Ortho-LoRA maintains full rank.}
\label{fig:singular_values}

\end{center}
\vspace{-1em}
\end{figure}

Linear dependence among the columns of LoRA matrices can compromise the effectiveness of the \textit{Vanilla T-LoRA} masking strategy, limiting the ability to exclude information effectively. Our analysis of LoRA weights in diffusion model personalization demonstrates that their effective rank is frequently much smaller than the specified rank (Figure~\ref{subfig:lora_sv}), demonstrating linear dependency between the matrix columns. Enforcing orthogonality in $A$ and $B$ matrices could address this, using an SVD-like architecture and regularization, as in AdaLoRA~\cite{adalora}:
\begin{equation} \label{eq:ada_lora}
\centering
\begin{split}
&\tilde{W} = W + BSA, \\
&L_{reg} = \lambda_{reg}(\| AA^T - I \|^2_F + \| B^TB - I \|^2_F)
\end{split}
\end{equation}
where $A \in \mathbb{R}^{r \times m}$ and $B \in \mathbb{R}^{n \times r}$ are normally initialized, and $S \in diag(\mathbb{R}^{r})$ is zero-initialized. AdaLoRA demonstrated that this setup requires approximately 10,000 iterations to achieve orthogonality, which is significantly higher than the 1,000-2,000 iterations typically needed for diffusion model customization. As a result, this regularization approach is not suitable for personalization tasks, making orthogonal initialization essential. Furthermore, initializing the \( S \) matrices with zeros considerably slows down the training process (see Section~\ref{subsec:initialization} for more details).

In summary, the task requires: (1) $A$ and $B$ to be orthogonal from the start, and (2) $S$ not to be zero-initialized. To address these, we use a \textit{LoRA trick} to revise LoRA weight initialization:

\begin{equation} \label{eq:lora_trick}
    \tilde{W} = \underbrace{W - BSA}_{\text{new weights}} + \underbrace{BSA}_{\text{LoRA}} = \hat{W} + BSA 
\end{equation}

This removes the need for zero initialization ($B=0$), enabling arbitrary weight initialization. As shown in Figure~\ref{subfig:tlora_scheme}, we initialize $A$ and $B$ using SVD~\cite{golub1971svd} factor matrices to enforce orthogonality: $A_{init} =V^T_r$, $B_{init} = U_r, S_{init}=S_r$, and $S_{init} = S_r$. We term this approach \textit{Ortho-LoRA} due to its orthogonal structure.

The choice of SVD for initialization is crucial. We examine six \textit{Ortho-LoRA} variants: using top, middle, and bottom singular components of original weights $W \in \mathbb{R}^{n \times m}$, and of a random matrix $R \in \mathbb{R}^{n \times m}$. Top-component initialization from original weights reduces to PISSA~\cite{pissa}, but we find it suboptimal for diffusion model customization. Initializing with the last SVD components of random matrix $R$ yields optimal results (see Section~\ref{subsec:initialization}). Figure~\ref{subfig:ortholora_sv} illustrates that, in contrast to LoRA, \textit{Ortho-LoRA maintains full rank throughout the entire training process without requiring any orthogonal regularization.}

\subsection{T-LoRA} \label{sec:tortholora}

Finally, we introduce the complete \textit{T-LoRA} framework (Figure~\ref{subfig:tlora_scheme}), combining \textit{Vanilla T-LoRA}'s timestep-dependent rank control with \textit{Ortho-LoRA}'s orthogonal initialization, resulting in a timestep-adaptive solution for diffusion model personalization:

\begin{equation} \label{eq:tlora}
\centering
\begin{split}
    &    \tilde{W} = W - B_{init}S_{init}M_tA_{init} + BSM_tA,
    \\
    &M_t = M_{r(t)} = \mathrm{diag}(\underbrace{1, 1, \dots, 1}_{r(t)}, \underbrace{0, 0, \dots, 0}_{r - r(t)})
\end{split}
\end{equation}
\noindent
where $A_{init} =V^T[-r:]$, $B_{init} = U[-r:]$ and $S_{init}=S[-r:]$ are the last SVD components of a random matrix $R = USV^T, R \sim \mathcal{N}(0, 1/r)$. The rank schedule follows $r(t) = \lfloor (r - r_{min}) \cdot (T-t)/T \rfloor + r_{min}$. We do not apply any orthogonality-enforcing regularization, as the Ortho-LoRA initialization inherently maintains the orthogonality of matrices throughout the entire training process.

%% file: sections/05_experiments_arxiv.tex
\begin{figure*}[t!]
  \centering
\includegraphics[width=\linewidth]{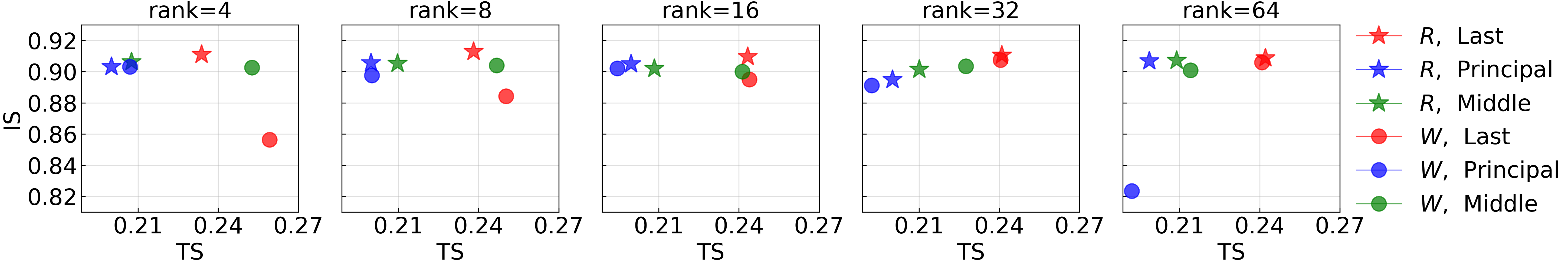}
\caption{Results for six \textit{Ortho-LoRA} initialization variants based on principal, middle, and last SVD components of the original weights \( W \), and a random matrix \( R \). Higher singular values correlate with overfitting, while too small values can slow down the training. Initialization with last singular values from \( R \) is optimal across most ranks.}
\label{fig:inits}
\end{figure*}

\section{Experiments} \label{sec:experiments}

\textbf{Dataset} We use 25 concepts from prior works~\citep{TI,DB,CD}, including pets, toys, interior objects, accessories, and more. Each concept is trained on a single, manually selected image with clear visibility. For evaluation, each is paired with 25 contextual prompts (appearance, position, background changes) and 6 complex prompts (e.g., background + accessorization). We generate 5 images per contextual prompt and 15 per base prompt (e.g., "a photo of $V^*$"), totaling 800 concept-prompt pairs. All experiments use all 25 concepts, except the Initialization Investigation, which uses 5 randomly sampled concepts to reduce compute.

\textbf{Evaluation Metric} To assess concept fidelity, we compute the average pairwise cosine similarity (IS) between CLIP ViT-B/32~\cite{clip} embeddings of real and generated images, following~\cite{TI}. Using the neutral prompt "a photo of $V^*$" ensures independence from appearance, position, and accessorization changes. Backgrounds are masked to reduce bias from training image reproduction. We also report DINO-IS, computed similarly with DINO~\cite{dino} embeddings. To evaluate prompt-image alignment (TS), we calculate the average cosine similarity between CLIP embeddings of the prompt and generated images.

\textbf{Experimental Setup} We investigated the performance of our method on two models: Stable Diffusion-XL~\cite{sdxl} and FLUX-1.dev~\cite{flux2024}. In all experiments models are fine-tuned with batch size 1, updating only the diffusion U-Net/DiT while keeping the text encoder fixed. Baselines follow their original setups.

\begin{figure}[t]
\centering
\includegraphics[width=\linewidth,trim={3em 1em 2.5em 3.5em}]{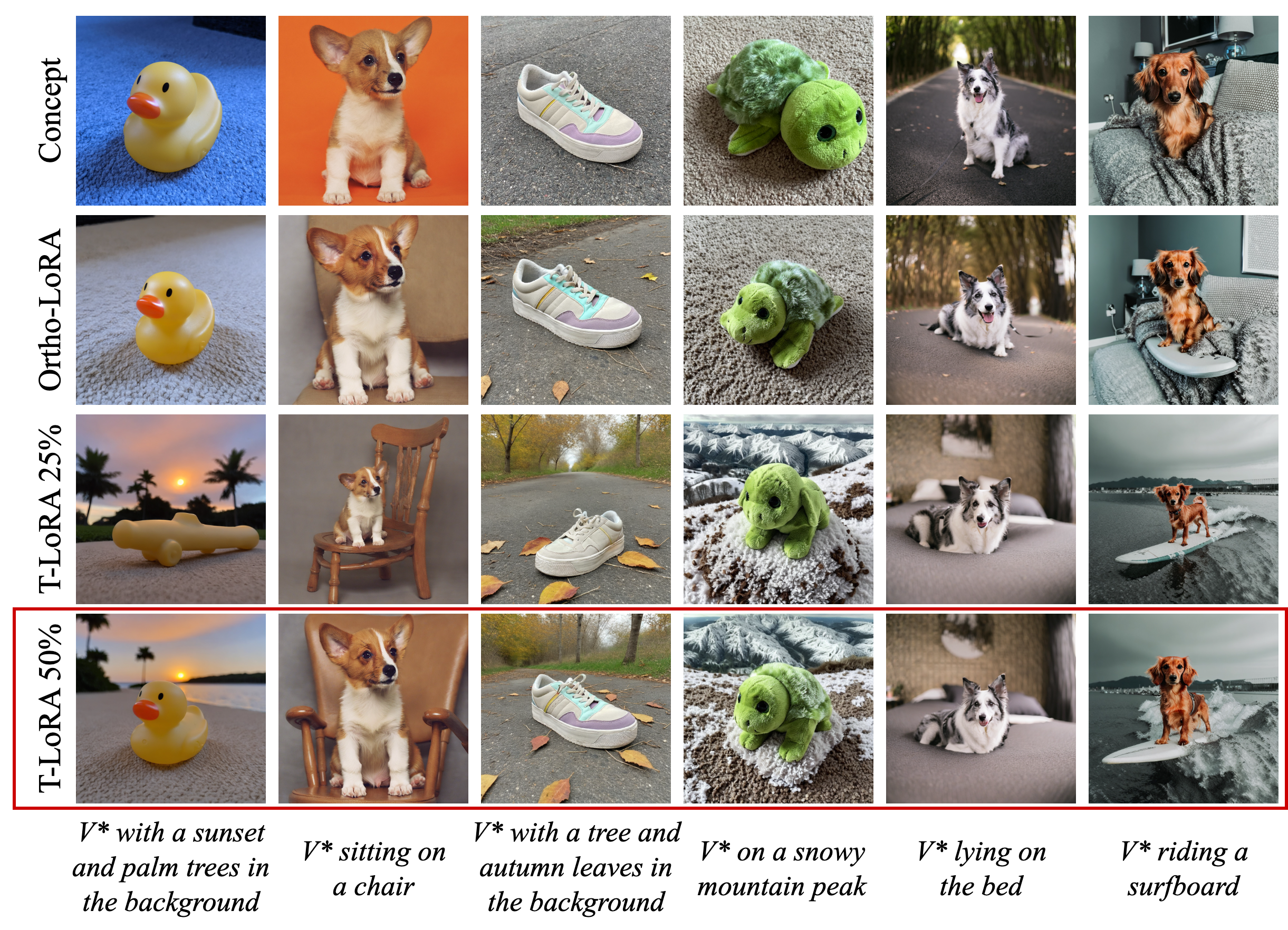}
\caption{\textbf{Ablation study.} Generation outputs for Ortho-LoRA and T-LoRA with \( r_{min} \) set to 25\% and 50\% of the full rank $r=64$. Ortho-LoRA generally exhibits poor alignment with the text and a high degree of reproduction of training images. In contrast, T-LoRA significantly enhances alignment with the text. Notably, T-LoRA at 50\% maintains high image fidelity, whereas T-LoRA at 25\% often struggles to accurately reproduce the concept.}
\label{fig:ablation_study}
\vspace{-1em}
\end{figure}

\subsection{T-LoRA Design Decisions} \label{subsec:initialization} 

\begin{table*}[t]

\centering
\caption{Image Similarity (IS) and Text Similarity (TS) for LoRA, Vanilla T-LoRA, and T-LoRA across dfferent ranks.}
\label{tab:ablations}

\begin{tabular}{lcccccccccc}
\toprule
\multirow{2}{*}{Methods} & \multicolumn{2}{c}{Rank = 4}    & \multicolumn{2}{c}{Rank = 8}    & \multicolumn{2}{c}{Rank = 16}   & \multicolumn{2}{c}{Rank = 32}                         & \multicolumn{2}{c}{Rank = 64}                                  \\
                         & IS             & TS             & IS             & TS             & IS             & TS             & IS    & TS    & IS             & TS    \\ \hline
LoRA                     & 0.890          & 0.250          & 0.897 & 0.249          & 0.900          & 0.243          & 0.901 & 0.238 & 0.901 & 0.232 \\
\textit{Vanilla T-LoRA}           & 0.894          & \textbf{0.259} & 0.892 & \textbf{0.261} & 0.902 & \underline{0.256}          & 0.904            & \underline{0.248}    & 0.902                     & \underline{0.240}                     \\
\textit{T-LoRA}                   & 0.899 & \underline{0.255}          & 0.897 & \underline{0.260}          & 0.897          & \textbf{0.260} & 0.899                     & \textbf{0.259}            & 0.900                              & \textbf{0.256} \\
\bottomrule
\end{tabular}

\end{table*}

\begin{table*}[t]
\vspace{0.6em}
\centering
\caption{
Image Similarity (IS) and Text Similarity (TS) for T-LoRA compared to the baseline models.}
\label{tab:methods_comparison}

\begin{tabular}{lccccccc}
\toprule
Metric & \textit{T-LoRA-64} & LoRA-64  & OFT-32 & OFT-16 & GSOFT-64 & GSOFT-32 & SVDiff \\ \hline
DINO-IS                      & 0.802  & \textbf{0.808} & 0.804    & 0.802    & \underline{0.806}      & 0.804      & 0.414    \\
IS                           & \underline{0.900}  & \textbf{0.901} & \textbf{0.901}    & 0.899    & \textbf{0.901}      & \textbf{0.901}      & 0.753    \\
TS                           & \underline{0.256}  & 0.232 & 0.247    & 0.212    & 0.247      & 0.212      & \textbf{0.295}   \\
\bottomrule
\end{tabular}
\end{table*}

\begin{table*}[h!]
\vspace{0.6em}
\centering
\caption{User study results of the pairwise comparison of T-LoRA versus other baselines.}
\label{tab:user_study}

\begin{tabular}{lcccccc}
\toprule
\multirow{2}{*}{Methods} & \multicolumn{2}{c}{Concept Preservation} & \multicolumn{2}{c}{Text Alignment} & \multicolumn{2}{c}{Overall Preference} \\  
                         & \textit{T-LoRA}              & Alternative        & \textit{T-LoRA}           & Alternative     & \textit{T-LoRA}              & Alternative      \\ \hline
\textit{Ortho-LoRA-64}              & \textbf{50.3}       & 49.7               & \textbf{58.5}    & 41.5            & \textbf{59.3}       & 40.7             \\
\textit{Vanilla T-LoRA-64}          & \textbf{51.7}       & 48.3               & \textbf{60.7}    & 39.3            & \textbf{60.3}       & 39.7             \\
LoRA-64                     & 39.3                & \textbf{60.7}      & \textbf{71.0}    & 29.0            & \textbf{67.3}       & 32.7             \\
OFT-32                  & \textbf{52.5}       & 47.5               & \textbf{58.3}    & 41.7            & \textbf{63.5}       & 36.5             \\
GSOFT-64                & 49.0                & \textbf{51.0}      & \textbf{61.5}    & 38.5            & \textbf{60.3}       & 39.7             \\
SVDiff                  & \textbf{90.1}       & 9.9                & 42.1             & \textbf{57.9}   & \textbf{55.9}       & 44.1   \\

\bottomrule
\end{tabular}

\end{table*}

\textbf{Analysis of Ortho-LoRA Initialization Strategies} We investigated six variants of the Ortho-LoRA initialization, which are based on the principal, middle, and last components of the original weights \( W \), and a random matrix \( R \). The IS and TS metrics for these setups are presented in Figure~\ref{fig:inits}. First, we observe that for all ranks and for both \( R \) and  \( W \) initializations the points in the TS metric are ordered according to the initialization singular values magnitude: the principal components initialization yields the lowest TS, followed by the middle components initialization, and the last components initialization yields the highest TS. This suggests that higher singular values are strongly correlated with overfitting. For ranks 4 and 8, initializing with the last SVD components of \( W \) turns out to be too close to zero and slows down the training process, whereas initializing with the last SVD components of \( R \) does not have this effect. Overall, initializing with the last SVD components from a random matrix \( R \) yields optimal results for most ranks, that is why we use it in all further experiments with Ortho-LoRA and T-LoRA.

\textbf{Selection of $r_{min}$} In Figure~\ref{fig:ablation_study}, we present generation examples for Ortho-LoRA and T-LoRA with \( r_{min} \) set to 25\% and 50\% of the full rank. Both T-LoRA variants significantly improve alignment with the text and enable a greater variety of positions and backgrounds for the concepts. As $r_{min}$ decreases, the generation becomes more flexible. However, while the 50\% performs well across most concepts, the 25\% is often too small, leading to reduced concept fidelity. Consequently, we use T-LoRA at 50\% in all subsequent experiments.

\subsection{Comparison with LoRA} 
In Table~\ref{tab:ablations}, we present image and text similarity for LoRA, Vanilla T-LoRA, and T-LoRA across various ranks on SD-XL. For all ranks, both Vanilla T-LoRA and T-LoRA demonstrate superior text similarity compared to LoRA, while maintaining same image similarity that differs from LoRA by only a third of a decimal place. At lower ranks, Vanilla T-LoRA and T-LoRA show similar performance; however, the performance improvement of T-LoRA becomes more pronounced as the rank increases. At low ranks, LoRA approaches full rank, which is why Vanilla T-LoRA performs comparably to T-LoRA. In contrast, as the ranks increase, the effectiveness of masking in Vanilla T-LoRA diminishes, while T-LoRA continues to demonstrate its full potential.

\subsection{Comparison with Baselines}
In addition to LoRA~\cite{lora}, we compare our T-LoRA with other lightweight customization methods, including OFT~\cite{ortogonal}, GSOFT~\cite{groupshouffle}, and SVDiff~\cite{svdiff}. The results on SD-XL are presented in Table~\ref{tab:methods_comparison}. 

T-LoRA achieves the best text similarity across all methods, except for SVDiff; however, SVDiff exhibits very low image similarity and often fails to accurately represent the concept. While LoRA demonstrates the highest image similarity, it also exhibits the most significant overfitting. Notably, our method's image similarity differs from LoRA's by only a third of a decimal place. 

Figure~\ref{fig:all_methods} showcases examples of generation for each method. T-LoRA provides greater flexibility in generation concerning position and background changes while accurately representing the concept.

\subsection{Multi-image Experiments} \label{app:multi_image}
In addition to the single-image experiments, we evaluate \textit{T-LoRA} against LoRA~\cite{lora} and OFT~\cite{ortogonal} in the multi-image diffusion model customization. In this setting, each concept is represented by multiple images featuring diverse backgrounds. Table~\ref{tab:multi_image_comparison} summarizes the results for experiments on SD-XL conducted with 1, 2, and 3 images. For \textit{T-LoRA} and LoRA we use $r=64$, and $n_{blocks} = 32$ for OFT as it showed the best results in single-image setup.

\textit{T-LoRA} consistently outperforms LoRA in text similarity across all image counts while achieving similar image similarity. Remarkably, \textit{T-LoRA} trained on one image surpasses LoRA trained on two or three images. Compared to OFT, \textit{T-LoRA} excels in the two-image scenario and performs similarly in the three-image scenario.

\begin{table}[h]
\centering
\resizebox{\linewidth}{!}{
\begin{tabular}{lcccccc}
\toprule
\multirow{2}{*}{Methods}               & \multicolumn{2}{c}{\# Images = 1} & \multicolumn{2}{c}{\# Images = 2} & \multicolumn{2}{c}{\# Images = 3} \\
                                       & IS          & TS                  & IS          & TS                  & IS          & TS                  \\ \hline
LoRA-64 & 0.901       & 0.232               & 0.900       & 0.245               & 0.902       & 0.251               \\
OFT-32                                 & 0.901       & \underline{0.247}               & 0.901       & \underline{0.261}               & 0.901       & \textbf{0.267}      \\
\textit{T-LoRA-64}                        & 0.900       & \textbf{0.256}      & 0.901       & \textbf{0.262}      & 0.900       & \underline{0.263}               \\ \bottomrule
\end{tabular}
}
\caption{Image Similarity (IS) and Text Similarity (TS) for multi-image Customization experiments.}
\label{tab:multi_image_comparison}
\end{table}

\FloatBarrier
\newpage
\begin{strip}
\centering
\includegraphics[width=0.75\linewidth]{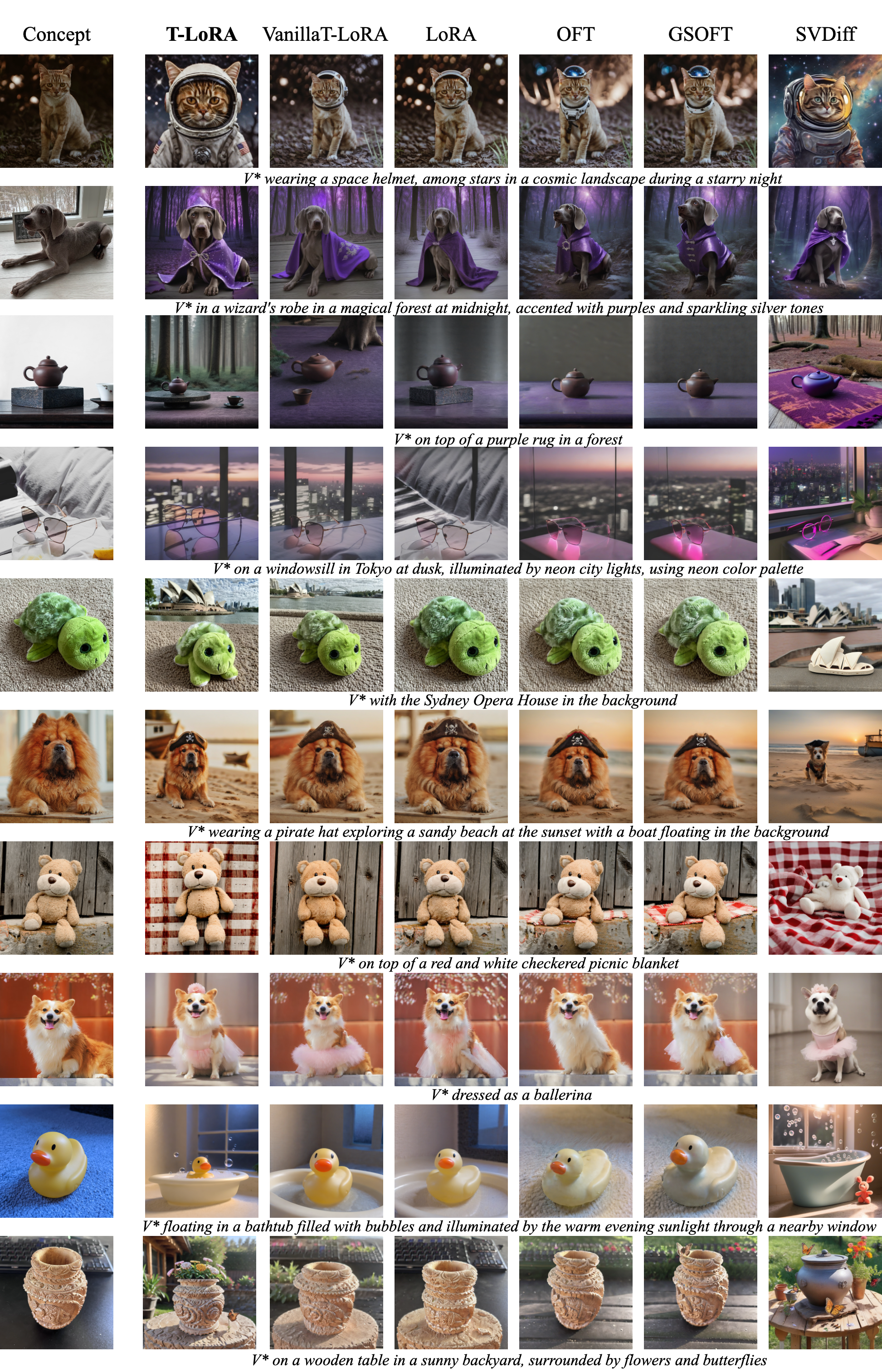}
\captionof{figure}{Generation examples for T-LoRA alongside other diffusion model customization baselines.}
\label{fig:all_methods}
\end{strip}

\FloatBarrier

\begin{strip}
\centering
\includegraphics[width=0.75\linewidth]{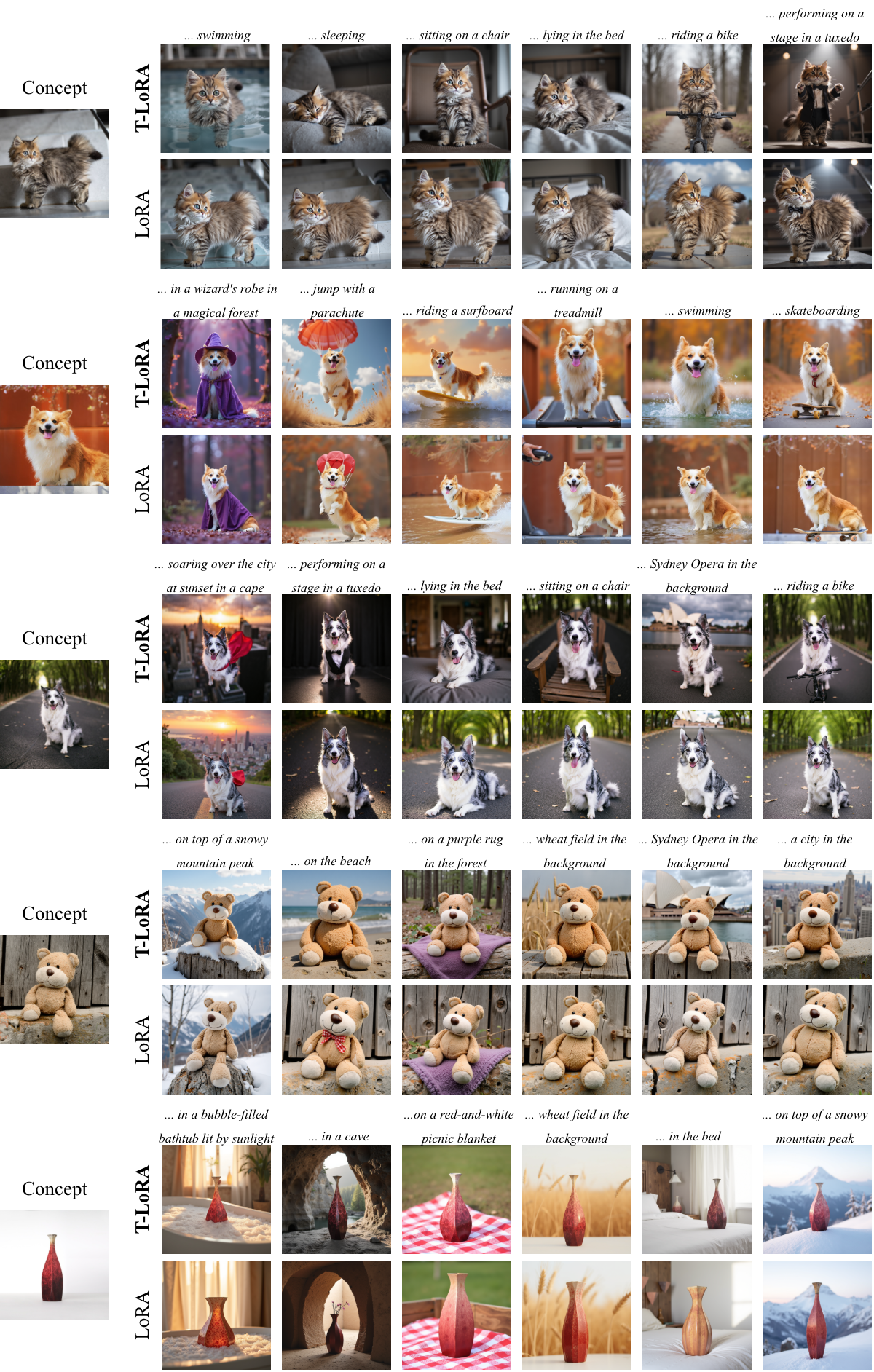}
\captionof{figure}{Generation examples for T-LoRA compared to standart LoRA on FLUX1.-dev.}
\label{fig:flux_all}
\end{strip}
\FloatBarrier

\subsection{User Study} \label{subsec:user_study}

We also conduct Human Evaluation to fully investigate our model's performance. Using an original image of the concept, a text prompt, and two generated images (one from T-LoRA and the other from an alternative method), we asked users to respond to the following questions: (1) "Which image more accurately represents the original concept?" to evaluate image similarity (2) "Which image aligns more closely with the text prompt?" to assess text similarity; and (3) "Which image overall demonstrates better alignment with the prompt and preserves the identity of the concept?" to evaluate the overall preference. For each pair of methods, we randomly generated 60 unique concept-prompt pairs. In total, we collect 1,800 human assessments across six method pairs.

Results in Table~\ref{tab:user_study} show that T-LoRA significantly outperforms others in text similarity and overall preference, while achieving comparable or superior image similarity to most methods. The exception is LoRA, which surpasses T-LoRA in image similarity due to its tendency to overfit and fully reproduce the original image. Despite this, T-LoRA maintains a strong overall impression, highlighting its balanced performance across criteria.

\subsection{FLUX-1.dev Experiments} \label{subsec:flux}

Finally, we investigated the performance of our model on the MM-DiT flow-based model, using FLUX-1.dev~\cite{flux2024}. For these experiments, we randomly selected 10 concepts from the full set: five live concepts and five objects. The prompt set was the same as in the SD-XL experiments.

\begin{figure}[t]
\centering
\includegraphics[width=0.49\linewidth]{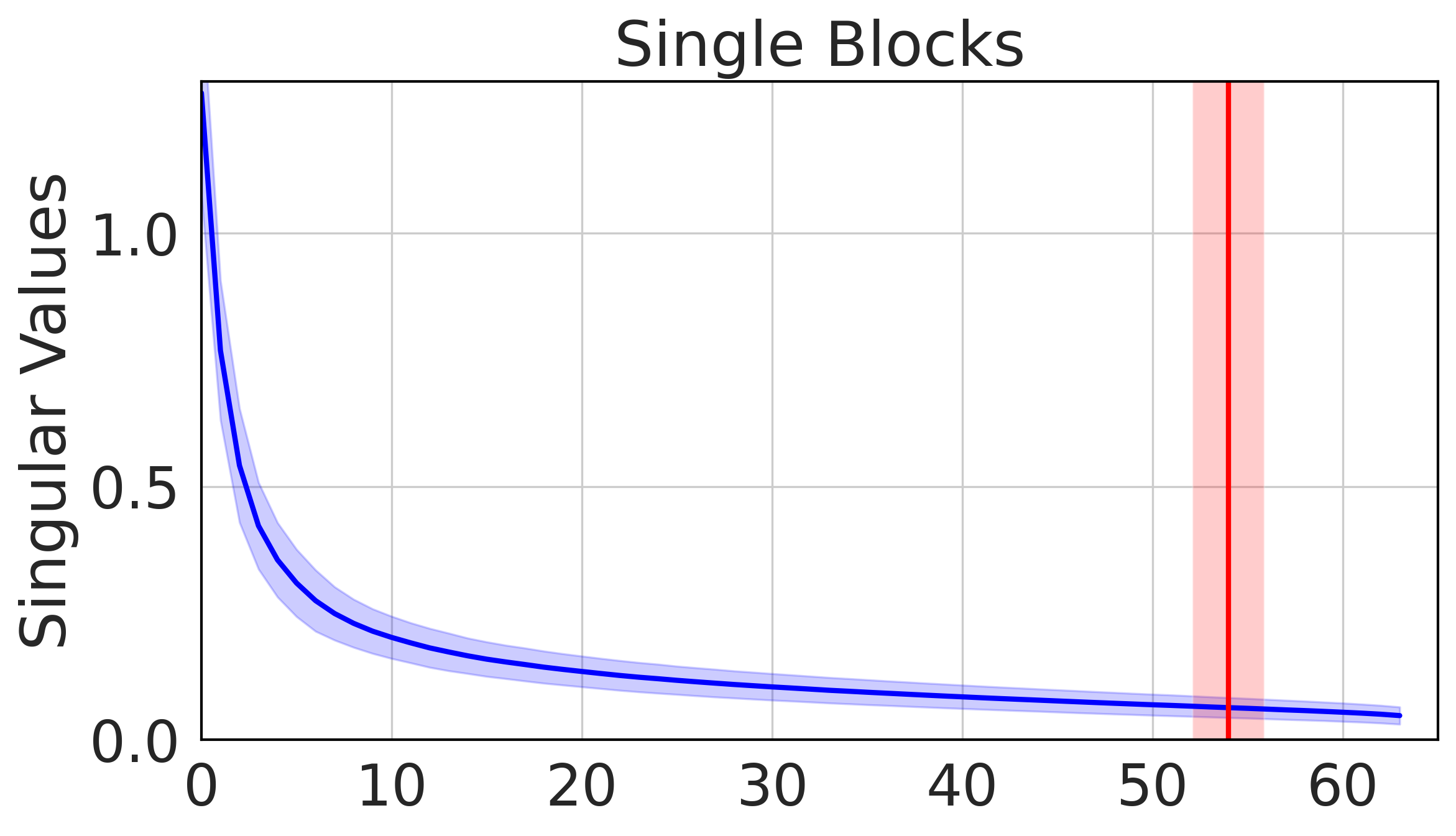}
\includegraphics[width=0.49\linewidth]{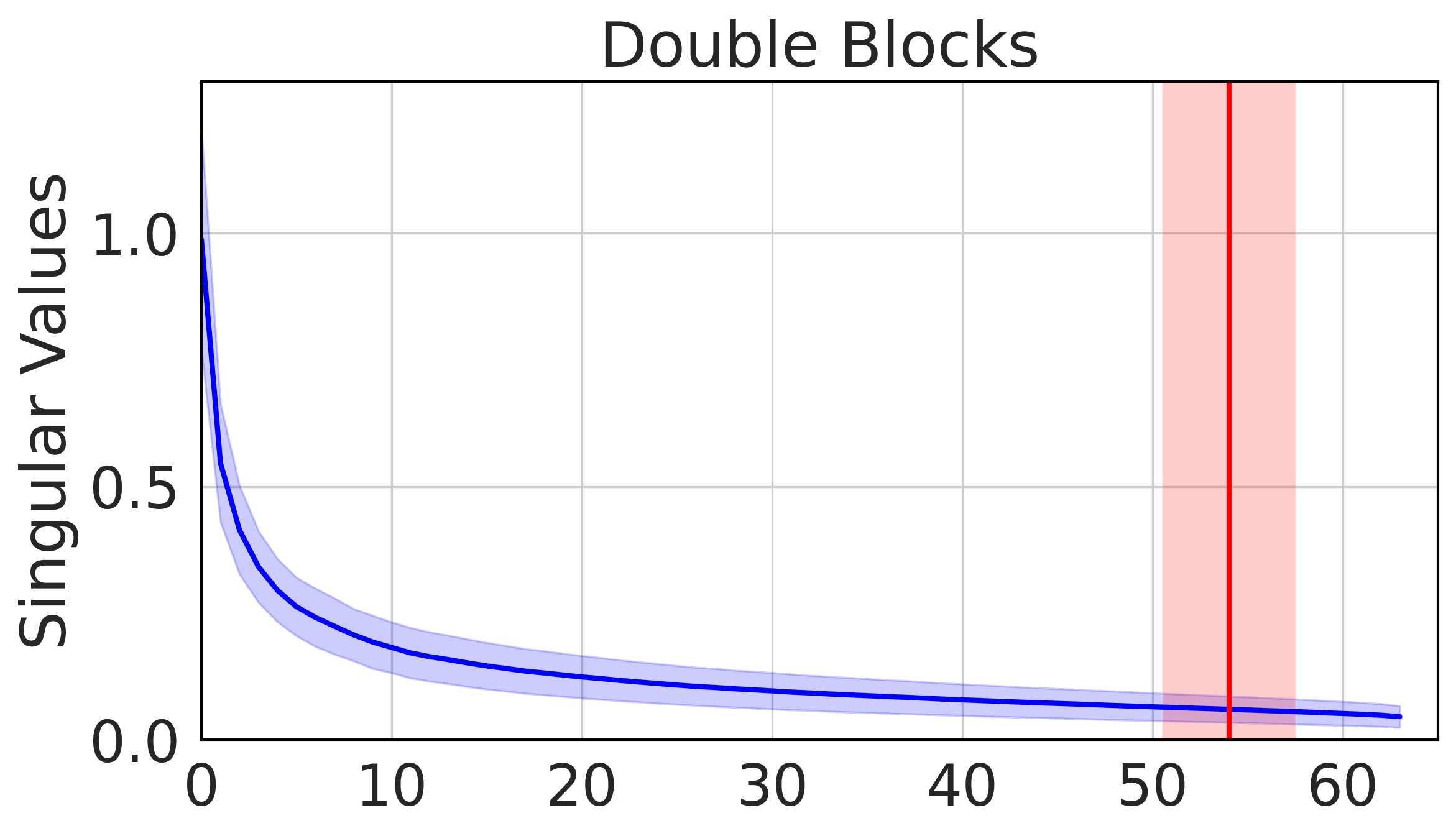}
\caption{Singular values of FLUX-1.dev LoRA B matrices with $r=64$ after 500 training steps. The red line marks the rank capturing \textbf{95\% of the total singular value sum}. Compared to SD-XL, all the singular values are non-zero and effective rank is near the full rank.}
\label{fig:flux_sv}
\end{figure}

We found that, unlike SD-XL, LoRA adapters trained on MM-DiT are full-rank (see Figure~\ref{fig:flux_sv}), making the additional orthogonal initialization unnecessary. Therefore, we used the Vanilla variant of T-LoRA for all FLUX-1.dev experiments.

The results are presented in Table~\ref{tab:flux} and Figure~\ref{fig:flux_all}. T-LoRA consistently achieves better prompt alignment and produces images with a more consistent ambiance than LoRA, as confirmed by both quantitative metrics and qualitative visual examples.

\begin{table}[t]

\centering
\caption{Image Similarity (IS) and Text Similarity (TS) for LoRA and T-LoRA across dfferent ranks on FLUX-1.-dev.}
\label{tab:flux}

\resizebox{\linewidth}{!}{
\begin{tabular}{lcccccccccc}
\toprule
\multirow{2}{*}{Methods} & \multicolumn{2}{c}{Rank = 4}    & \multicolumn{2}{c}{Rank = 16}   & \multicolumn{2}{c}{Rank = 64}                                  \\
                         & IS             & TS              & IS             & TS                & IS             & TS    \\ \hline
LoRA                     & 0.890          & 0.263          & \textbf{0.905}          & 0.264      & 0.884 & 0.247 \\

\textit{T-LoRA}                   & \textbf{0.908} & \textbf{0.268} & 0.903          & \textbf{0.280}             & \textbf{0.888}       & \textbf{0.280} \\
\bottomrule
\end{tabular}
}

\end{table}

%% file: sections/06_conclusion.tex
\section{Conclusion} \label{sec:conclusion}

This paper addressed the challenge of personalizing diffusion models using a single concept image, where overfitting and limited generative diversity are prevalent. We introduced \textit{T-LoRA}, a Timestep-Dependent Low-Rank Adaptation framework featuring (1) a rank masking strategy to regulate training across diffusion timesteps and (2) \textit{Ortho-LoRA}, an orthogonal weight initialization technique to enhance effective rank utilization. Extensive experiments show that \textit{T-LoRA} outperforms prior methods, balancing concept fidelity and text alignment in data-limited settings. Our results highlight the potential of timestep-aware adaptation and orthogonality for advancing text-to-image generation and related creative tasks.

%% file: sections/appendix_extended.tex
%%%%%%%%%%%%%%%%%%%%%%%%%%%%%%%%%%%%%%%%%%%%%%%%%%%%%%%%%%%%%%%%%%%%%%%%%%%%%%%
%%%%%%%%%%%%%%%%%%%%%%%%%%%%%%%%%%%%%%%%%%%%%%%%%%%%%%%%%%%%%%%%%%%%%%%%%%%%%%%
% APPENDIX
%%%%%%%%%%%%%%%%%%%%%%%%%%%%%%%%%%%%%%%%%%%%%%%%%%%%%%%%%%%%%%%%%%%%%%%%%%%%%%%
%%%%%%%%%%%%%%%%%%%%%%%%%%%%%%%%%%%%%%%%%%%%%%%%%%%%%%%%%%%%%%%%%%%%%%%%%%%%%%%
\onecolumn
\newpage
\appendix

\section{Limitations} \label{app:limitations}
Firstly, our method introduces an additional hyperparameter, \( r_{\text{min}} \), to the standard LoRA framework. While we demonstrate that setting \( r_{\text{min}} \) to 50\% of the full rank \( r \) performs well in most cases, the optimal choice may vary across different concepts. Furthermore, a lower \( r_{\text{min}} \) can contribute to more diverse generation, but it may also necessitate longer training times or a more complex choice of \( r(t) \), such as a non-linear approach, which we do not address in this work. Lastly, the SVD initialization of the weights introduces a slight increase in time and computational overhead compared to the standard LoRA initialization.

\section{Training Details} \label{app:training-details}
\subsection{SD-XL}
All the models, except SVDiff are trained using Adam optimizer with $\text{batch size} = 1$, $\text{learning rate} = 1e-4$, $\text{betas} = (0.9, 0.999)$ and $\text{weight decay} = 1e-4$.  
All the experiments were conducted using a single GPU H-100 per each model customization.

\textbf{LoRA} We implement LoRA, Vanilla T-LoRA, Ortho-LoRA and T-LoRA based on the \url{https://github.com/huggingface/diffusers}. For all ranks we train LoRA and Vanilla T-LoRA for 500 training steps and 800 trainings steps for Ortho-LoRA and T-LoRA

\textbf{OFT} We implement OFT with PEFT library~\cite{peft}. We train all OFT models for 800 training steps.

\textbf{GSOFT} We use offitial repo \url{https://github.com/Skonor/group_and_shuffle} for GSOFT implementation. We train all GSOFT models for 800 training steps.

\textbf{SVDiff} We implement the method based on \url{https://github.com/mkshing/svdiff-pytorch}. The model for all concepts were trained for $1600$ training steps using Adam optimizer with $\text{batch size} = 1$, $\text{learning rate} = 1e-3$, $\text{learning rate 1d} = 1e-6$, $\text{betas} = (0.9, 0.999)$, $\text{epsilon} = 1e\!-\!8$, and $\text{weight decay} = 0.01$. 

\subsection{FLUX-1.dev}
All the models are trained using prodigy optimizer with $\text{batch size} = 1$, $\text{learning rate} = 1$. We trained LoRA and T-LoRA for $400$ and $500$ training steps, respectively.

\begin{figure*}[h!]
\centering
\includegraphics[width=0.92\linewidth]{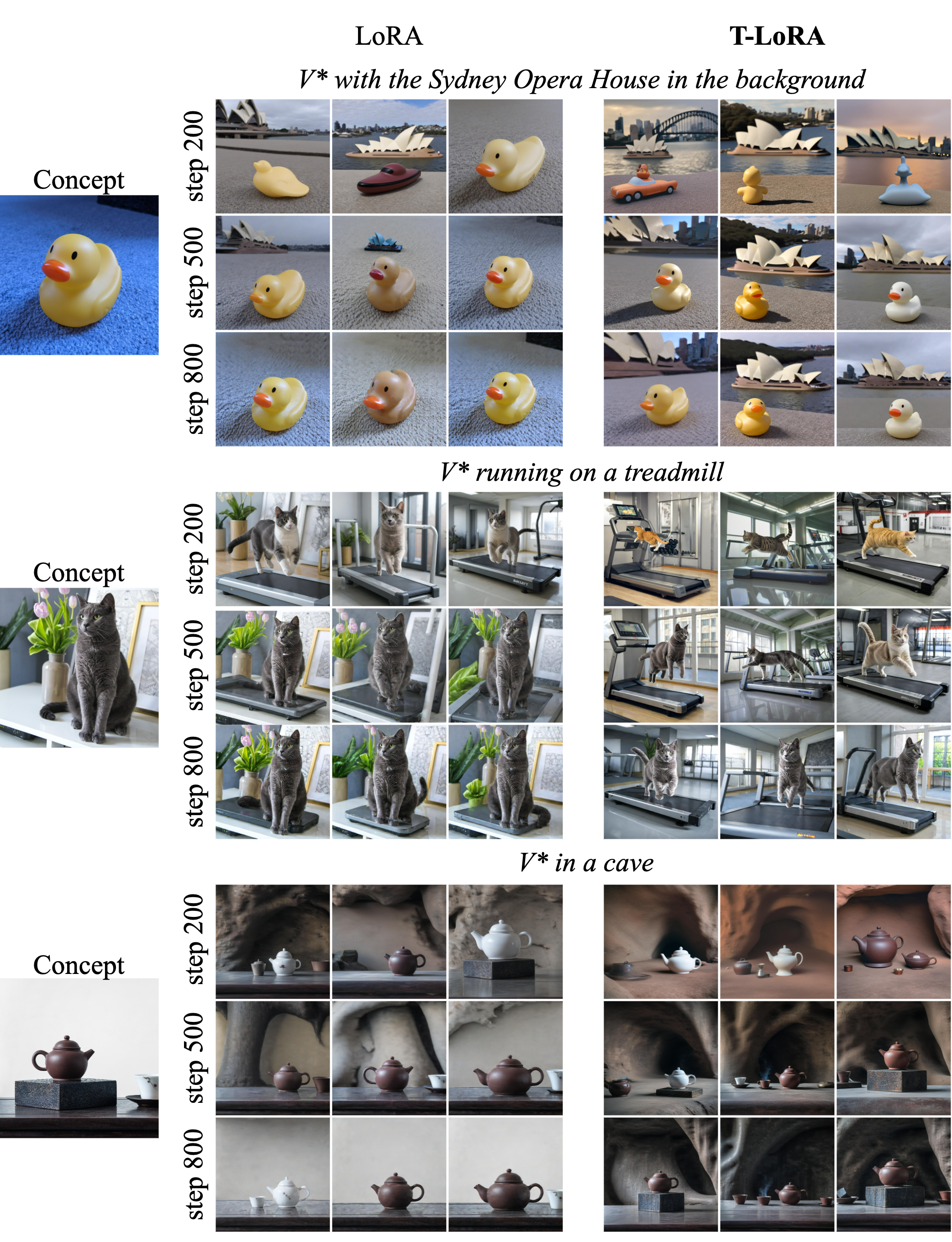}
\caption{Generation examples for LoRA and T-LoRA $r=64$ on the different training steps.}
\label{fig:td64}
\end{figure*}

\begin{figure*}[h!]
\centering
\includegraphics[width=0.92\linewidth]{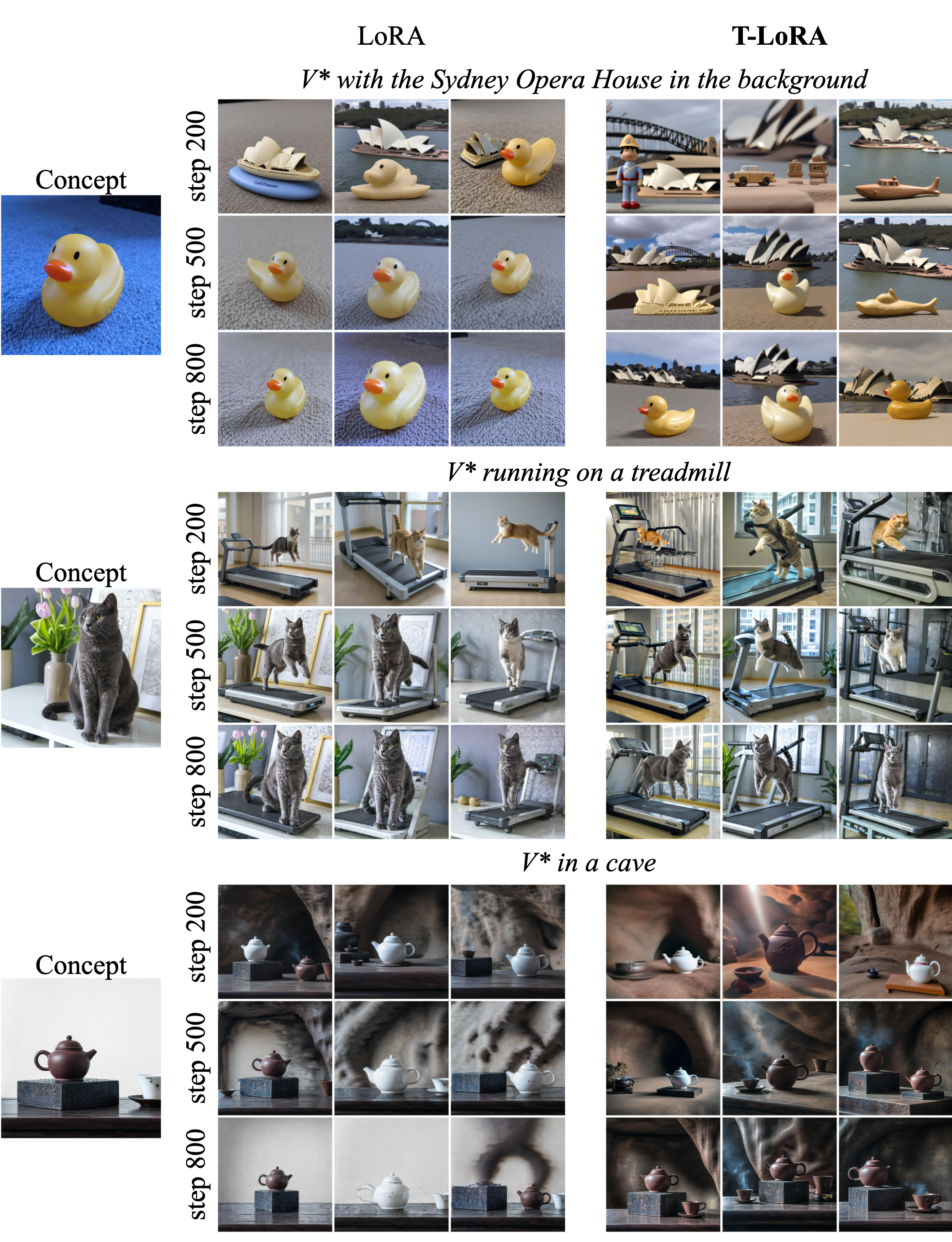}
\caption{Generation examples for LoRA and T-LoRA $r=16$ on the different training steps.}
\label{fig:td16}
\end{figure*}

\begin{figure*}[h!]
\centering
\includegraphics[width=0.92\linewidth]{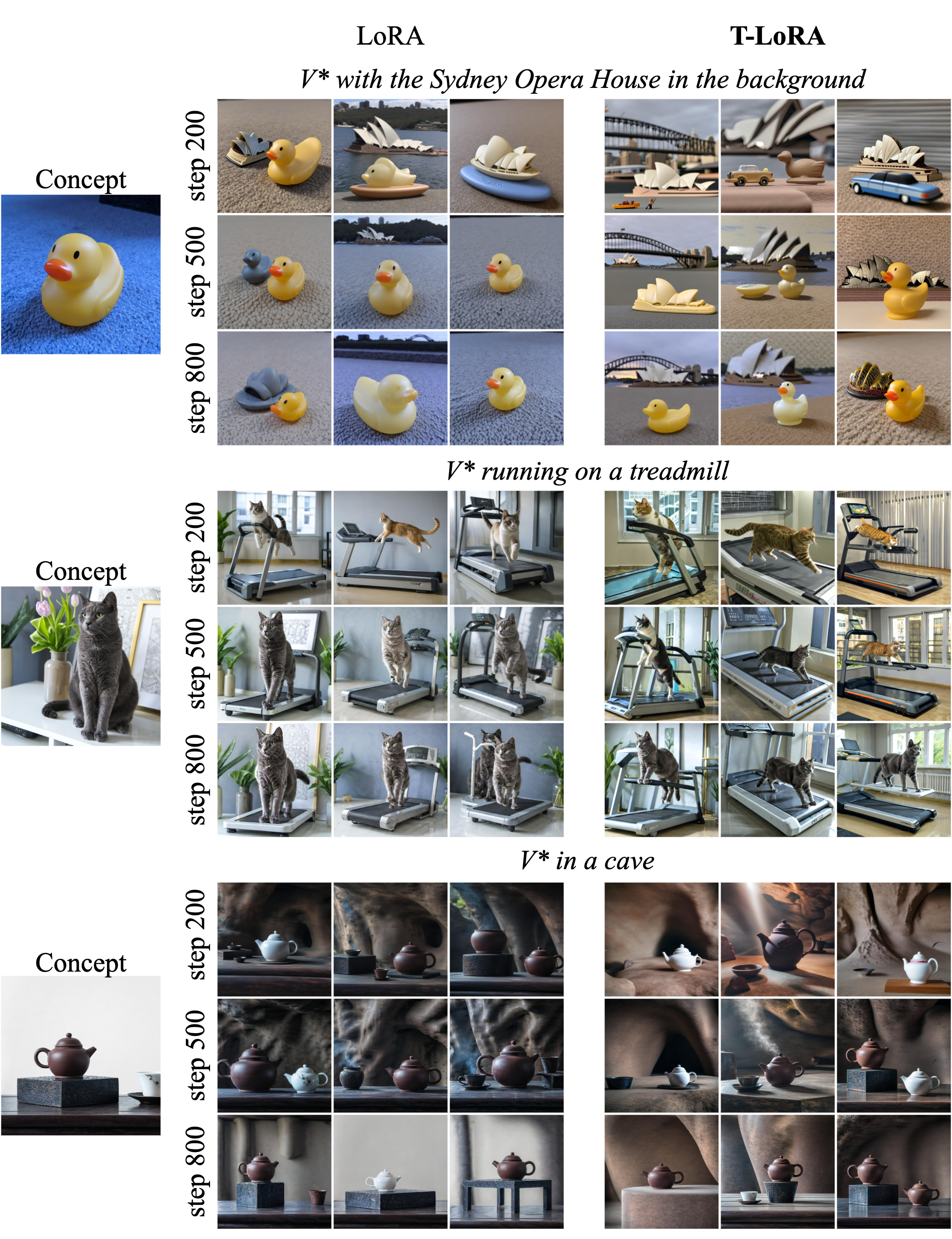}
\caption{Generation examples for LoRA and T-LoRA $r=4$ on the different training steps.}
\label{fig:td4}
\end{figure*}

\section{Training Dynamics} \label{app:training_dynamics}

In Figures~\ref{fig:td64}, \ref{fig:td16}, and \ref{fig:td4}, we present generation examples for LoRA and T-LoRA at different training steps across various ranks ($r = [64, 16, 4]$). A common observation during LoRA fine-tuning is that it tends to overfit, producing examples with poor prompt alignment while it has not yet fully learned the concept. This effect is particularly evident in the 200 training steps for the duck toy (Figure~\ref{fig:td64}), where the third image lacks an appropriate background, while the first two images fail to accurately represent the concept. Similarly, in the example with the cat (Figure~\ref{fig:td64}), the generation at 200 steps includes a flower from the training set, while the cat itself is not represented correctly. In contrast, T-LoRA typically requires slightly longer training to grasp the concept, but it learns more smoothly and exhibits significantly less overfitting in terms of both position and background. This observation is consistent across the other ranks as well.
\newpage
\section{Prompts} \label{app:prompts}

We use the following prompts in our experiments:

\vspace{1em}
\adjustbox{max width=1.0\textwidth}{
\begin{lstlisting}[numbers=none]
live_set = [
    'a chocolate-colored V*',
    'a red V*',
    'a purple V*',
    'a brightly lit V*',
    'a snowy V*',

    'a V* in a purple wizard outfit',
    'a V* in a doctor outfit',
    'a V* in a superhero suit',
    'a V* dressed as a ballerina',
    'a V* wearing a green bow tie',

    'a V* with the Eiffel Tower in the background',
    'a V* with a mountain in the background',
    'a V* with a tree and autumn leaves in the background',
    'a V* on top of green grass with sunflowers around it',
    'a V* with a sunset and palm trees in the background',
    'a V* with the Sydney Opera House in the background',

    'a V* lying in the bed',
    'a V* sitting on the chair',
    'a V* swimming ',
    'a V* sleeping ',
    'a V* skateboarding',

    'a V* riding a surfboard',
    'a V* jumping with a parachute',
    'a V* running on a treadmill',
    'a V* riding a bike',

    'a V* in a chief outfit in a nostalgic kitchen filled with vintage furniture and scattered biscuit',
    'a V* dressed in a superhero cape, soaring through the skies above a bustling city during a sunset',
    'a V* dressed wearing a tuxedo performing on a stage in the spotlight',
    "a V* in a wizard's robe in a magical forest at midnight, accented with purples and sparkling silver tones",
    'a V* wearing a space helmet, among stars in a cosmic landscape during a starry night',
    'a V* wearing a pirate hat exploring a sandy beach at the sunset with a boat floating in the background',
]

object_set = [
    'a chocolate-colored V*',
    'a V* made of gold',
    'a brightly lit V*',
    'a red V*',
    'a purple V*',

    'a V* on top of a red and white checkered picnic blanket',
    'a V* with the Eiffel Tower in the background',
    'a V* with a mountain in the background',
    'a V* with a tree and autumn leaves in the background',
    'a V* on top of green grass with sunflowers around it',
    'a V* with a sunset and palm trees in the background',
    'a V* with the Sydney Opera House in the background',
    'a V* in the snow',
    'a V* on the beach',
    'a V* on a cobblestone street',
    
    'a V* on top of a wooden floor',
    'a V* with a city in the background',
    'a V* with a blue house in the background',
    'a V* on top of a purple rug in a forest',
    'a V* with a wheat field in the background',

    'a V* in the bed',
    'a V* on the chair',
    'a V* on top of a snowy mountain peak',
    'a V* with a sand castle in the background',
    'a V* in a cave',

    'a V* on a windowsill in Tokyo at dusk, illuminated by neon city lights, using neon color palette',
    'a V* on a sofa in a cozy living room, rendered in warm tones',
    'a V* on a wooden table in a sunny backyard, surrounded by flowers and butterflies',
    'a V* floating in a bathtub filled with bubbles and illuminated by the warm glow of evening sunlight filtering through a nearby window',
    'a V* floating among stars in a cosmic landscape during a starry night with a spacecraft in the background',
    'a V* on a sandy beach next to the sand castle at the sunset with a floaing boat in the background',
]
\end{lstlisting}
}